# Morphological classifiers

É.O. Rodrigues [a,*], A. Conci [b], P. Liatsis [c]

[a] *Institute of Science and Technology, Universidade Federal de Itajuba (UNIFEI), Minas Gerais, Brazil*
[b] *Department of Computer Science, Universidade Federal Fluminense, Niterói - Rio de Janeiro, Brazil*
[c] *Department of Computer Science, Khalifa University of Science and Technology, Abu Dhabi, United Arab Emirates*

## A B S T R A C T

This work proposes a new type of classifier called Morphological Classifier (MC). MCs aggregate concepts from mathematical morphology and supervised learning. The outcomes of this aggregation are classifiers that may preserve shape characteristics of classes, subject to the choice of a stopping criterion and structuring element. MCs are fundamentally based on set theory, and their classification model can be a mathematical set itself. Two types of morphological classifiers are proposed in the current work, namely, Morphological k-NN (MkNN) and Morphological Dilation Classifier (MDC), which demonstrate the feasibility of the approach. This work provides evidence regarding the advantages of MCs, e.g., very fast classification times as well as competitive accuracy rates. The performance of MkNN and MDC was tested using $p$-dimensional datasets. MCs tied or outperformed 14 well established classifiers in 5 out of 8 datasets. In all occasions, the obtained accuracies were higher than the average accuracy obtained with all classifiers. Moreover, the proposed implementations utilize the power of the Graphics Processing Units (GPUs) to speed up processing.

## 1. Introduction

Supervised learning consists of analysing labelled instances in order to generate a classification model capable of producing labels for unlabelled instances. Supervised learning is well established and vastly employed in fields such as computer science and engineering, and in a great number of sub areas, such as robotics [1], computer vision [2,3], data mining, knowledge discovery [4], health care [5–8] and many others [9].

Classification algorithms are built upon disctinct concepts and theories. Some are based on decision-trees, others on neural networks, Bayes theorem, mathematical functions, rules, regressions, etc. The performance of each classifier varies according to the used dataset. According to the *no free lunch* theorem, there is no one model that works best for every optimization problem. The assumptions of a great model built for a specific problem may not hold for another problem [10]. Usually, multiple models are evaluated in order to determine the best for a particular problem.

In this work, we propose the use of mathematical morphology (MM) in the context of classification. Mathematical Morphology [11,12] is commonly employed to process and analyse geometric structures in images and is based on set theory. We introduce the theory of morphological classifiers and propose two paradigms for this new classification approach, one inspired mostly by k-Nearest Neighbours (k-NN) [13] and morphology and the other by morphological dilations [14].

The work is organized as follows. In the next section, we perform a state-of-the-art review in supervised learning and mathematical morphology. In Sections 3 and 4, we define the main proposal of this work and present the MkNN and MDC algorithms. Section 5 is concerned with experiments and performance comparison to other state-of-the-art approaches. Finally, in the last section, we provide the conclusions and propose avenues for future work.

## 2. Literature review

A brief introduction to supervised learning, followed by an overview of the various types of classification algorithms is introduced in this section. Next, we introduce the essential operations of Mathematical Morphology and provide examples of its application.

### 2.1. Classification algorithms

Lazy algorithms [15] generate classification models in real time. That is, the algorithm generates a model right before classifying unlabelled instances. Their main weakness stems from that fact, i.e., generating new models when unlabelled instances are encountered requires rework. However, this results into a significant strength, i.e., there is no requirement to store a model in memory.

* Corresponding author.
*E-mail addresses:* erickr@id.uff.br (É.O. Rodrigues), aconci@ic.uff.br (A. Conci), panos.liatsis@ku.ac.ae (P. Liatsis).

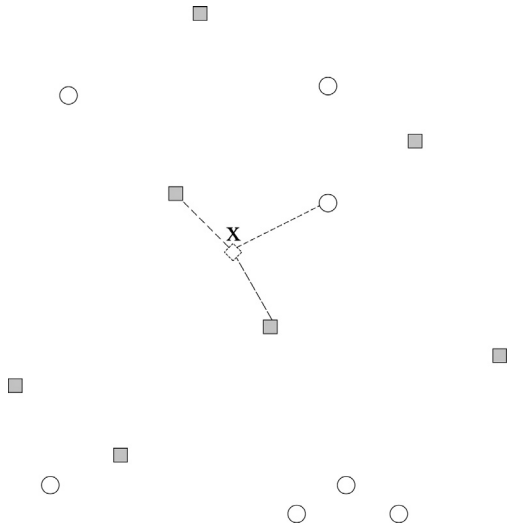

**Fig. 1.** Example of a k-NN algorithm in a 2D plane. *x* is the unlabelled instance and *k* is set to 3. In this case *x* would be classified as a square.

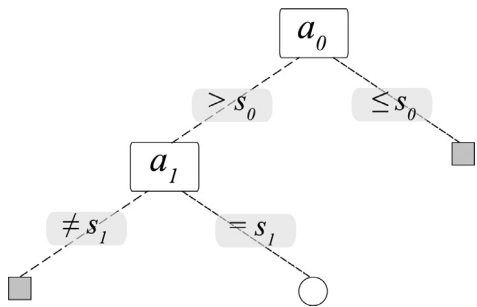

**Fig. 2.** A decision tree model with two nodes (two split criterions). The leaves of the trees represent the class of the unlabelled instance.

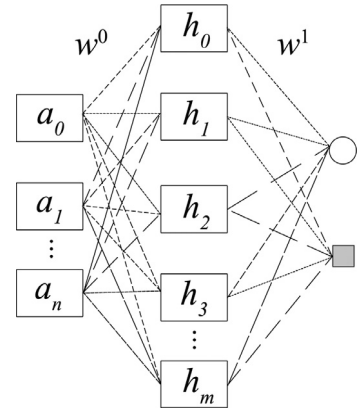

**Fig. 3.** A simple neural network with one hidden layer consisting of *m* units. The hidden units receive a weighted combination of the input attributes, which passes through activation functions, finally resulting in the final classification. $w^0$ are the weights between the hidden and input units, while $w^1$ are the weights between the outputs and hidden units.

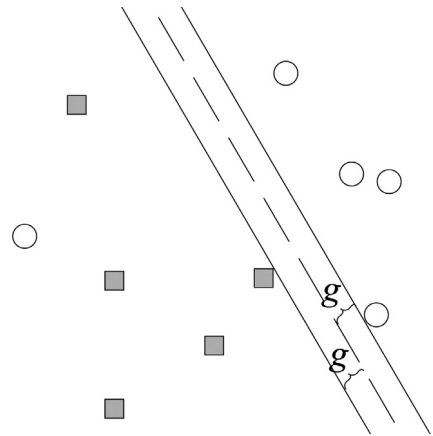

**Fig. 4.** A possible 2D SVM model. In this case, the training instances are separated by the dashed line so that if an unlabelled instance falls on the left/right side of the line, it would be classified as a square/circle, respectively.

k-NN is one of the most popular lazy algorithms and probably the simplest one. It requires the *k* nearest neighbours of an unlabelled instance to be examined in order to determine its class. The predominant class among the *k* neighbours of an unlabelled instance is chosen as the target class. k-NN is a robust algorithm in terms of accuracy, yet it does not deal well with noise in some specific cases and for some values of *k*. One such occasion is when a small number of instances of one class is surrounded by several instances of other classes. A simple k-NN with $k = 3$ is depicted in Fig. 1. The unlabelled instance is checked against the 3 nearest labelled instances and is classified as the most frequently encountered class that, in this case, is the square class.

Decision tree based classification relies on decision tree models. Essentially, these algorithms generate one or more trees that, in turn, contain several nodes. Each node split is usually associated to a formulated Boolean criterion, which depends on one or more attributes of the dataset. For instance, $a_i > s$, where $a_i$ is the attribute at index *i* and *s* is a constant. The split governs the flow of the decision, and the leaves of the trees represent the various classes. A popular decision tree algorithm is C4.5 [16,17] (J48 in Weka [18]). Random Forests [19] is also frequently employed in classification. It generates several decision trees that, when combined together, compose an ensemble learning method [20]. Fig. 2 illustrates a possible decision tree based classification model.

Another type of classificier is neural networks. Neural networks are often categorized as function-based classifiers. They associate input patterns to output patterns by using the representation of intermediate layers, activation functions and weights that connect the various nodes, mimicking the organization of the brain. Two popular neural networks are the Radial Basis Function Network (RBF Network) [21,22] and the Multilayer Perceptron (MLP) [23,24]. Fig. 3 shows a feedforward neural network with *n* input units, *m* hidden units and two outputs. The outputs of the hidden units are the weighted sums of the input attributes after they pass through the activation functions of these units. Finally, the outputs of this neural network are calculated as the weight sums of the input from the hidden units after they pass through the activation function of the output units.

Support Vector Machines (SVM), on the other hand, are examples of function-based classifiers. In summary, SVM traces hyperplanes that segregate the data, and uses the created clusters or segregations to predict classes. In its basic form, SVM works in a straightforward fashion. Fig. 4 illustrates a possible solution, shown by the dashed line, which separates the data optimally. Optimality in this case refers to higher accuracy and margin, i.e., gap size *g*. Since this problem is 2-dimensional, the hyper-plane is a simple line. Algorithms that derive from SVM such as Sequential Minimal Optimization (SMO) [25] also bear the same optimization objective.

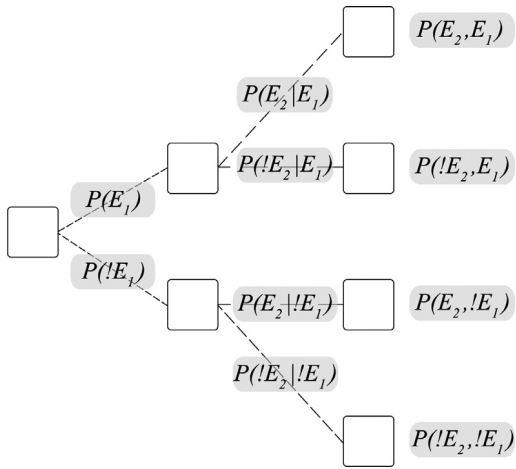

**Fig. 5.** A simple Bayes network. The joint probabilities on the right side of the figure indicate the probability of an instance belonging to a certain class.

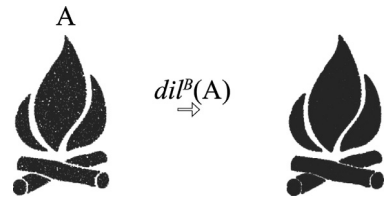

**Fig. 7.** A binary image dilation by structuring element *B*.

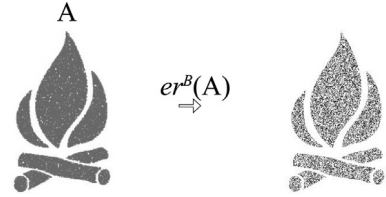

**Fig. 8.** A binary image erosion by structuring element *B*.

**Fig. 6.** A decision table that predicts the class of unlabelled instances based on the rules $r_0$, $r_1$ and $r_2$. If $r_0$ and $r_1$ are true, the instance is probably a circle.

Some of the previously described algorithms can be thought of as probabilistic models. That is, some decision tree algorithms, for instance, output a probability on the correctness of classification of a certain node, or the respective path of the tree that a given instance followed. These probabilities can be computed in various ways and therefore, these algorithms could be considered probabilistic. However, some classifiers are based on conditional probabilities, more specifically, on Bayes Theorem [26]. These are explicitly categorized as probabilistic. Here, we refer to them as Bayes-based algorithms, with the popular algorithms being the Naive Bayes [27,28] and Bayes Net [6,29]. In Bayes theorem, $P(E_i)$ denotes the probability of a given event $E_i$ being true. The event could be, for instance, the probability of an attribute being equal, greater, lower or different than a certain value. $P(E_a|E_b)$ denotes the conditional probability, which is the probability of $E_a$ being true given that $E_b$ is true. A simple Bayes net is shown in Fig. 5. The joint probabilities at the right of the figure indicate the probability of the analysed instance belonging to a certain class, given the described events. The exclamation marks indicate the *NOT* operator.

Lazy, decision tree, function, and Bayes classifiers are probably the most commonly used classifers in the state-of-the-art. However, there is another category of classifiers that often produces very good results, yet it is often disregarded. These are rule-based classifiers such as Decision Table [30] and Conjutive Rule [18]. A decision table can be extraordinarily simple. Such table can be constructed using the default rule mapping to the majority class. In general, decision tables are relatively easy to understand as generated rules are often simple and constructed using perfectly understandable logical constraints. The construction process involves heuristical search of feature subsets for rule finding, as well as cross-validation [30].

Fig. 6 shows a decision table model. The first variables $r_0$, $r_1$ and $r_2$ represent rules based on training instances. In the second column, if rules $r_0$ and $r_1$ are true, then the table classifies the instance as a circle. In the fifth column, on the other hand, if an unlabelled instance has rules $r_0$ and $r_1$ equal to false and $r_2$ true, then it is classified as a square. The rule itself can be a simple comparison split of an attribute (e.g., if an attribute value is higher than a constant), as in decision tree models, or more complex associations.

### 2.2. Mathematical morphology (MM)

Mathematical morphology [12,31] is concerned with geometric structures present in images. The morphological analysis of images is performed using set theory [32]. The structuring element that is used in most mathematical morphology operations is defined by a mathematical set of elements that represents its pixels. Dilations and erosions are the two principal and fundamental operations in MM. A dilation of a binary image $A = \{a_1, a_2, \ldots, a_n\}$, containing $n$ foreground pixels, by a structuring element $B$, is given by [14]:

$$dil^B(A) = \bigcup_{b \in B} A_b \quad (1)$$

where $A_b$ represents the elements of *A* translated by *b*. In other words, dilation is given by the union of $A + b$ for every element $b$ in *B*. Each element represents the position of a pixel. Fig. 7 illustrates a dilation of a binary image *A* by a cross-shaped structuring element $B = \{(0, 0), (0, 1), (0, -1), (1, 0), (-1, 0)\}$.

The dilation operation sets the intensity of any background pixels via the superposition of the structuring element on each of the foreground pixels of the binary image to the foreground value. In this specific case, since the spatial extent of the structuring element is fairly limited, it spreads the foreground pixels that already exist to their neighbouring pixels, thus filling the gaps of the original image.

Erosion is the opposite operation to dilation, and is defined as [14]:

$$er^B(A) = \bigcap_{b \in B} A_{-b} \quad (2)$$

Fig. 8 shows the erosion of an image *A* by the same cross-shaped structuring element *B*. Erosion erases pixels of an image respecting the structuring element. In this specific case, since the spatial extent of the structuring element is fairly limited, it erodes the pixels near the original ones, so that holes or noise in the image become more evident.

Mathematical morphology involves a wide variety of operations such as opening, closing, top-hat, reconstruction and others. These operations are derived from the essential operations of dilation and erosion. For instance, opening is a combination of erosions and dilations, while closing is the opposite, a combination of dilations and erosions, respectively. Top-hat is the result of subtracting the original image from its opening [33]. Morphological reconstruction, on the other hand, reconstructs the shapes of the image using seed points, dilations and a mask [34].

In general, mathematical morphology is often employed in the literature for geometric feature filtering, e.g., to segment structures [35], extracting and enhancing blood vessels [36,37], etc. Liu et al., [38] use grey-level mathematical morphology operations and a genetic algorithm to create a thresholding method for images. Furthermore, it is also used to remove noise from images [39,40]. The noise removal potential is evidenced in Fig. 7, where a great deal of noise from the original image is removed with a simple dilation. The closing operation is also used to remove noise from images [41].

The closest that mathematical morphology comes to classification in the existing literature is by providing image descriptors for further classification [42,43]. The power of mathematical morphology has not yet been explored in the context of classification models. In this work, we provide a theoretical foundation for mathematical morphology-based classifiers, and propose two such models.

## 3. Theoretical framework

The essence of the approach is to expand pattern sets of labelled instances such that the expansion covers the entire feature space while preserving the morphology of the original clusters of labelled instances. We commence a process that subsets are expanded using set operations from their original state $i = 0$, to their final state, $i = I_l$, for each class $l$. For this, we focus on data-driven classification functions, based on the $p$-dimensional shape of the original class clusters. The expansion in a specific direction may be terminated when a suitable termination criterion is reached. This criterion could be, for instance, the classification error $P_{err} > t_{err}$, where $t_{err}$ is a predefined misclassification threshold.

We consider a class of labelled patterns represented by a number of disjoint clusters as a multiset. A multiset is a generalization of the set concept, which allows the presence of repeated instances. For example, {1, 2, 2} is a different multiset in comparison to {1, 2} although they are the same set. Thus, given a training dataset $X = \{x_1, \ldots, x_n\}$ that is a multiset containing instances such that $x \in \mathbb{C}$, where $\mathbb{C} \subset \mathbb{R}^p$, $p$ being the number of attributes of the instances, with corresponding labels $c(x) \in \{1, \ldots, L\}$, an unlabelled instance x is assigned a label $l \in \{1, \ldots, L\}$, such that:

$$\hat{c}(x) = l : x \in A_l^{I_l} \quad (3)$$

subject to

$$\bigcup_{l=1}^{L} A_l^{I_l} = \mathbb{C} \quad (4)$$

where $I_l$ corresponds to the final iteration for class $l$ and $A_l^{I_l}$ is the expanded set partition of instances belonging to class $l$, which consists of $J_l$ disjoint subsets $a_{l,j}^{I_l}$ such that:

$$A_l^{I_l} = \bigcup_{j=1}^{J_l} a_{l,j}^{I_l} \quad (5)$$

In what follows, we describe the methodology for obtaining the expanded set partition $A_l^{I_l}$. Let us assume a set of pattern points $A_l^0$ belonging to class $l$, with index 0 referring to the original set partition of training samples:

$$A_l^0 = \{x \in X \ : \ c(x) = l\} \quad (6)$$

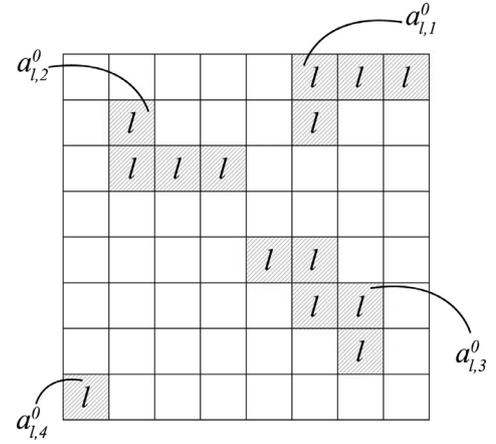

**Fig. 9.** Partition $A_l^0$ consisting of 4 disjoint subsets at iteration 0.

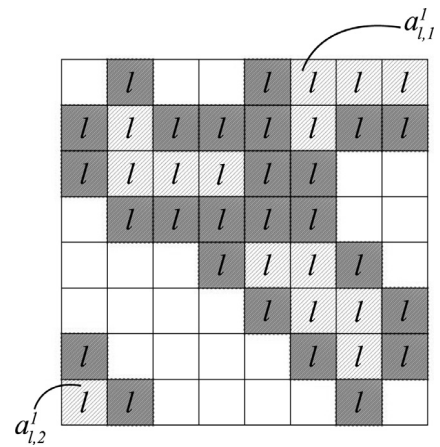

**Fig. 10.** Set partition $A_l^1$ consisting of 2 disjoint subsets at iteration 1, where the dark grey shaded pixels represent the elements that are included in the set at iteration 1.

The original set partition of samples belonging to class $l$, $A_l^0$, is represented by the union of the disjoint subsets $a_{l,j}^0$. An example of this in two dimensions is shown in Fig. 9. Here, the class set partition $A_l^0$ is the union of $\{a_{l,1}^0, a_{l,2}^0, a_{l,3}^0, a_{l,4}^0\}$.

Next, we consider the expanded set partition $A_l^{i+1}$, corresponding to the application of a set operator $f$ at iteration $i$:

$$\begin{aligned} A_l^{i+1} &= f(A_l^i) \\ &\vdots \quad \vdots \\ A_l^1 &= f(A_l^0) \end{aligned} \quad (7)$$

where $f(A)$ is a function that expands set $A$.

Therefore, given $I_l$ iterations of the expanded class set partitions, we have $L$ equivalence classes as follows:

$$\begin{array}{llll} \{A_1^0, & A_1^1, & \ldots, & A_1^{I_1}\} & \textit{Equivalence class } 1 \\ \{A_2^0, & A_2^1, & \ldots, & A_2^{I_2}\} & \textit{Equivalence class } 2 \\ \vdots & \vdots & & \vdots \\ \{A_L^0, & A_L^1, & \ldots, & A_L^{I_L}\} & \textit{Equivalence class } L \end{array} \quad (8)$$

A candidate set morphology-preserving operation $f$ based on mathematical morphology is the dilation [14]:

$$f(A_l^i) = dil^B(A_l^i) \quad (9)$$

by a suitable structuring element B. The result of this for iteration 1 is shown in Fig. 10, assuming that iteration

0 is shown in Fig. 9, where in this specific case $B = \{(0, 0), (-1, 0), (1, 0), (0, -1), (0, 1)\}$.

In a dilator classifier, all $A_l^0$ can be expanded simultaneously, where intersections of labels are erased when they occur. The result of the dilation of $L$ classes until they cover the entire domain of $\mathbb{C}$ is a classification model.

Operation $f$ may be any other operation apart from strict dilations. For instance, it could be a combination of morphological operations such as dilations and erosions, a pseudo-random or random function operating on the set partition, etc.

In the case of $p$-dimensional set partitions $A_l^0$, the patterns are represented in a $p$-dimensional grid or hypercube. Each original set partition $A_l^0$ is represented by a unique class label $l$, and an appropriate set operator, such as a dilation, shown in Eq. 9, is applied to produce the expanded set partition $A_l^i$ at iteration $i$. The set expansion process is controlled by a convergence function, which monitors a performance measure, e.g., classification error. The evolution of a class set partition at iteration $i$ is terminated, according to the convergence function, and the membership of the class set partition is reset to that at iteration $i - 1$. In what follows, we provide the definition of a Morphological Classifier (MC):

**Definition 1.** Morphological Classifier (MC): Assume a partition $\mathbb{C}$ of $\mathbb{R}^p$ and $L$ set partitions $A_l^{I_l}$ such that:

$$\bigcup_{l=1}^{L} A_l^{I_l} = \mathbb{C} \tag{10}$$

where $I_l$ corresponds to the final iteration for class $l$, and each of the set partitions is the result of morphology-preserving set expansion operations.

A morphological classifier $M$ is a $p$-dimensional grid or hypercube, where the value of each element refers to the predicted label of an unclassified pattern $y$.

$$\hat{c}(y) = l : y \in A_l^{I_l}, l \in \{1, \ldots, L\} \tag{11}$$

The evolution of set partitions $A_l^{I_l}$ is governed by a convergence function $\phi()$.

Given a validation dataset with $m$ instances, $Y = \{y_1, \ldots, y_m\} \subset \mathbb{C}$ and a multiset with corresponding labels for this dataset $\{c(y_1), \ldots, c(y_m)\} \in \{1, \ldots, L\}$, its accuracy rate is given by:

$$Accuracy = \frac{|\{y \in Y : \hat{c}(y) = c(y)\}|}{|Y|} \tag{12}$$

where $\hat{c}(y)$ is the predicted class of instance $y$ and $|Y|$ stands for the cardinality of set $Y$.

Theorem 3.1 illustrates a possible methodology for the expansion of the set partitions. The fundamental concept of this approach is that it is possible to exploit redundancy, i.e., the complement of the union of $L - 1$ set partitions representing one of the set partitions, which reduces memory and computational requirements.

However, with $L$ set partitions, $L$ options for the complement set partition are possible, which leads us to an optimization problem. Theorem 3.1 proves that the selection of the complement partition affects accuracy.

**Theorem 3.1.** *Impact of complement set partition $A_q$ on classification performance:*

*Consider the $L$ expanded set partitions $A_l^i$ at iteration $i$, $l \in \{1, \ldots, L\}$, then*

$$A_q^i = \mathbb{C} \setminus \bigcup_{l=1}^{L} \begin{cases} A_l^i, & if\ l \neq q \\ \emptyset, & otherwise \end{cases} \tag{13}$$

*subject to*

$$\mathbb{C} = \bigcup_{l=1}^{L} A_l^i, \mathbb{C} \subset \mathbb{R}^p \tag{14}$$

*There is a set partition $A_q^{I_q}$, $q \in \{1, \ldots, L\}$, at final iteration $I_l$, that maximizes accuracy for a specific classification problem.*

**Proof.** Eq. 13 introduces the use of the complement of a set partition in the expansion process, which reduces memory and processing time requirements. For instance, in binary classification, the partitions are $A_1^i$ and $A_2^i$. The expansion is performed in one of the two partitions, while the other corresponds to the complement of the first partition.

Assuming that $X_l$ is a multiset of instances with labels equal to $l$, then the accuracy rate for the binary case is given by:

$$Accuracy = \frac{|X_1 \cap A_1^{I_1}| + |X_2 \cap A_2^{I_2}|}{|X|} \tag{15}$$

Assuming $A_q^{I_q} = A_2^{I_2}$, then $|X_2 \cap A_2^{I_2}|$ is equivalent to $|X_2 \setminus A_1^{I_1}|$, as $A_2^{I_2}$ is the complement set partition ($A_2^{I_2} = \mathbb{C} \setminus A_1^{I_1}$). Thus, the accuracy for $A_q^{I_q} = A_2^{I_2}$ is given by:

$$Accuracy_{A_q^{I_q} = A_2^{I_2}} = \frac{|X_1 \cap A_1^{I_1}| + |X_2 \setminus A_1^{I_1}|}{|X|} \tag{16}$$

The previous accuracy concerns to $A_q^{I_q} = A_2^{I_2}$. In the other case, where $A_q^{I_q} = A_1^{I_1}$, instead of expanding $A_1^{I_1}$, $A_2^{I_2}$ is expanded, which provides the following accuracy rate:

$$Accuracy_{A_q^{I_q} = A_1^{I_1}} = \frac{|X_2 \cap A_2^{I_2}| + |X_1 \setminus A_2^{I_2}|}{|X|} \tag{17}$$

Thus, we want to prove that the choice of $A_q^i$ has an impact on the accuracy. For this to be true, we assume that Eq. 18 is true and derive contradiction. If this does not hold for at least a single case, then the theorem is proved.

$$\begin{aligned} Accuracy_{A_q^{I_q} = A_2^{I_2}} &= Accuracy_{A_q^{I_q} = A_1^{I_1}} \\ \frac{|X_1 \cap A_1^{I_1}| + |X_2 \setminus A_1^{I_1}|}{|X|} &= \frac{|X_2 \cap A_2^{I_2}| + |X_1 \setminus A_2^{I_2}|}{|X|} \end{aligned} \tag{18}$$

Let us assume a similar case for both set accuracy computations, where $A_1^{I_1}$ and $A_2^{I_2}$, respectively, are expanded until domain $\mathbb{C}$ is fully covered, such that $A_1^{I_1} = A_2^{I_2} = \mathbb{C}$. This implies $X_2 \setminus A_1^{I_1} = \emptyset$, $X_1 \setminus A_2^{I_2} = \emptyset$, $X_1 \cap A_1^{I_1} = X_1$ as well as $X_2 \cap A_2^{I_2} = X_2$. Therefore:

$$\begin{aligned} \frac{|X_1 \cap A_1^{I_1}| + |X_2 \setminus A_1^{I_1}|}{|X|} &= \frac{|X_2 \cap A_2^{I_2}| + |X_1 \setminus A_2^{I_2}|}{|X|} \\ |X_1 \cap A_1^{I_1}| + |X_2 \setminus A_1^{I_1}| &= |X_2 \cap A_2^{I_2}| + |X_1 \setminus A_2^{I_2}| \\ |X_1 \cap A_1^{I_1}| + |\emptyset| &= |X_2 \cap A_2^{I_2}| + |\emptyset| \\ |X_1 \cap A_1^{I_1}| &= |X_2 \cap A_2^{I_2}| \\ |X_1| &= |X_2| \end{aligned} \tag{19}$$

$|X_1| = |X_2|$ is a contradiction as $X_1$ and $X_2$ can be different multisets. Therefore, the accuracy can be affected by the choice of the complement set partition. □

## 4. Methods

In this section, we propose two classification algorithms based on morphology. The first algorithm consists of an adaptation of the k-NN algorithm to the proposed morphological framework. The second algorithm is a mathematical morphology classifier based on simple dilations.

Before introducing the proposed classifiers, we need to consider that the input dataset is preprocessed so that the classification space resembles a discrete grid in order to speed up computations and for ease of implementation. Given a dataset $X$, each position of the discrete grid $G$ contains a number of repeated instances in $X$ and is defined by:

$$G_{round(\nu \circ x)} = |\{x \in X\}| \tag{20}$$

where $\nu$ is a precision vector $\nu = \langle \nu_1, \ldots, \nu_p \rangle$ and $round(\nu \circ x)$ is also a $p$-dimensional vector. $\nu \circ x$ is the Hadamard product of these two vectors:

$$\nu \circ x = \langle \nu_1 x_1, \ldots, \nu_k x_k, \ldots, \nu_p x_p \rangle \tag{21}$$

where in this specific case $x_k$ stands for the $k^{th}$ attribute of $x$.

The grid is a subset of natural numbers representing $\mathbb{C}$, which increases the processing performance of the algorithms and facilitates data compression.

### 4.1. Morphological k-NN (MkNN)

In this section, we present an algorithm similar to k-NN, however conforming to the theoretical definitions of Section 3. This algorithm generates a classification model, which does not require re-training and thus, is no longer a lazy classifier. However, as in the traditional k-NN classifier, an unlabelled instance $y$ has its surroundings examined until the algorithm checks at least $k$ other instances. Finally, it classifies $y$ as the majority class of the visited neighbouring instances.

Formally, given an unlabelled instance $y$, the neighbouring instances are visited so that at iteration $i$, labelled instances that are at a distance $\leq i$ from $y$ are visited. If at any iteration $i$, the number of visited labelled instances is $\geq k$, where $k$ is an input parameter as in the k-NN algorithm, then the unlabelled instance $y$ is said to be of the same class as the majority of the visited labelled neighbouring instances. The $f$ operation in this case is given by:

$$A_l^{i+1} = f(A_l^i) = A_l^i \cup \{y \in \mathbb{C} : \Omega(A_l^i, y) = l\} \tag{22}$$

The algorithm proceeds by analysing and examining the nearest labelled instances. For a single unlabelled position $y$ in grid $G$, the surrounding labelled instances $x$ are examined, as shown in function $\Omega()$:

$$\Omega(A_l^i, y) = mode(\{x \in X : d(x, y) \leq i\}) \tag{23}$$

where $d$ is a distance function and $mode()$ returns the most frequent label in the multiset. This process is repeated for every possible unlabelled position in the grid before the algorithm reaches convergence, i.e., it terminates when $k$ nearest instances have been visited. The convergence function for class $l$ is then given by:

$$\phi(A_l^i) = \begin{cases} 1, & \text{if } \sum_{y \in \mathbb{C}} \phi_a(y) = |\mathbb{C}| \\ 0, & \text{otherwise} \end{cases} \tag{24}$$

subject to

$$\phi_a(y) = |\{x \in X : d(x, y) \leq i\}| \geq k \tag{25}$$

where $|\mathbb{C}|$ indicates the total size of the subset $\mathbb{C}$ or grid $G$, i.e., the total amount of elements in the grid. Finally, the labelling function of the MkNN algorithm conforms to Eq. 11. The pseudo-code for constructing this classification model is shown in Algorithm 1.

### 4.2. Morphological dilator classifier (MDC)

Morphological Dilator Classifiers inherit more of the properties of mathematical morphology in contrast to MkNN. This algorithm dilates positions in grid $G$ until a termination criterion is reached. The $f$ operation for MDC is governed by a rule $\beta$ that indicates the orientation in which the instances should be dilated. Function $f$ is then given by:

$$A_l^{i+1} = f(A_l^i) = dil(A_l^i, b(d, i, \beta)) \tag{26}$$

**Data**: $k$ and $\gamma$ are input parameters, where $k$ stands for how many neighbouring instances should be visited, $\gamma$ stands for a weight applied to the central grid position and $\sigma$ represents the maximal number of iterations. $T_1, \ldots, T_L$ are counters for the classes, $X$ is the training dataset, $A_l^0$ respects Eq. 6 and $c(x)$ returns the class label of instance x.

1 **begin**
2   **for** each possible y position in grid G **do**
3     $k_{aux} \leftarrow 0; i \leftarrow 0;$
4     **for** each class $l$, such that $1 \leq l \leq L$ **do** $T_l \leftarrow 0;$
5     **if** $y \in X$ **then** $T_{c(y)} \leftarrow \gamma;$
6     **while** $k_{aux} < k$ **do**
7       **for** each instance $x \in X : d(y, x) = i$ **do**
8         $T_{c(x)} \leftarrow T_{c(x)} + 1;$
9         $k_{aux} \leftarrow k_{aux} + 1;$
10       $i \leftarrow i + 1;$
11       **if** $\sigma < i$ **then** break;
12     $maxl \leftarrow$ receives the $l$ that maximizes $T_l;$
13     $A_{maxl}^{i+1} \leftarrow \left( \bigcup_{i2=0}^{i2<i+1} A_{maxl}^{i2} \right) \cup \{y\};$
14   Return $A_1^{l_1}, \ldots, A_L^{l_L};$

**Algorithm 1:** Construction of the MkNN classification model.

where the operator $dil$ refers to a $p$-dimensional dilation of the expanded set partition $A_l^i$ at iteration $i$ by the $p$-dimensional structuring element $b$. Structuring element $b$ is a set whose structure is defined by the choice of distance metric $d$, distance measure $i$ and orientation factor $\beta$.

Examples of expansion rules for $\beta$, which alter the orientation of the structuring element and hence the expansion of the associated set partition for the case of two dimensions, are:

$$\beta = \begin{cases} e_1 \geq -|e_2| & : \text{to the left} \\ e_2 \geq |e_1| & : \text{to the bottom} \\ e_1 \leq |e_2| & : \text{to the right} \\ e_2 \leq -|e_1| & : \text{to the top} \end{cases} \tag{27}$$

where $e_1$ represents the $x$ coordinate of $e$ and $e_2$ its $y$ coordinate. Combinations of these rules are also possible. Different expansions and $p$-dimensional rules can also be applied. For the MDC implementation in this work, we consider these 4 different orientations and any combination of them, resulting in a total of 16 distinct growing patterns.

At low level overview, the $f$ function MDC is given by:

$$A_l^{i+1} = f(A_l^i) = A_l^i \cup \{y \in \mathbb{C} : d(x, y) = i, x \in A_l^i, \beta\} \tag{28}$$

Besides, the convergence function of MDC is given by:

$$\phi(A_l^i) = \begin{cases} 1, & \text{if } P_{err}(A_l^i) < t_{err} \vee i \geq \sigma \\ 0, & \text{otherwise} \end{cases} \tag{29}$$

where $P_{err}$ is a classification error measure and $t_{err}$ is an error threshold, indicating the maximum accepted error prior to the classifier terminating training. $\sigma$ forces expansions to terminate if $i$ exceeds $\sigma$. $\vee$ represents the Boolean operation OR. The individual error $P_{instErr}(y)$ of an instance $y$ is given by:

$$P_{instErr}(y) = \begin{cases} 1, & \text{if } \tau T_l < T_u, u \in \{1, \ldots, L\}, u \neq l \\ 0, & \text{otherwise} \end{cases} \tag{30}$$

where $T_l$ is a counter of neighbouring instances of $y$ that have been engulfed by the expansion and belong to class $l$. $\tau$ is a scaling factor that, such as in GRASP [44], allows the expansion to continue even if $T_l < T_u$. The total error of the set partition is the sum of

$P_{instErr}$ such that:

$$P_{err}(A_l^i) = \sum_{y}^{y \in A_l^i} P_{instErr}(y) \tag{31}$$

The pseudo-code for generating the MDC classification model is shown in Algorithm 2.

**Data**: $\gamma$, $\tau$, $\beta$ and $\sigma$ are input parameters. $\gamma$ stands for a weight applied to the central instance, $\tau$ controls the convergence as shown in Eq. 30, $\beta$ a orientation or expansion altering rule, as shown in Eq. 27, and $\sigma$ is a threshold for the number of partition set expansions. $T_1, \ldots, T_L$ are counters for the classes $1 \ldots L$, $X$ is the training dataset and $A_l^0$ respects Eq. 6. $q$ indicates the complement set partition.

1 **begin**
2   **for** *each class l, such that* $1 \leq l \leq L, l \neq q$ **do**
3     $i \leftarrow 0$;
4     **for** *each instance* $e \in A_l^i$ **do**
5       **for** *each class l2, such that* $1 \leq 2l \leq L$ **do**
        $T_{l2} \leftarrow 0$;
6       $T_{c(e)} \leftarrow \gamma$;
7       **while** $P_{instErr}(e) = 0$ *and* $i < \sigma$ **do**
8         **for** *each instance* $x \in X : d(e, x) \leq i$ **do**
9           $T_{c(x)} \leftarrow T_{c(x)} + 1$;
10         **if** $P_{instErr}(e) = 0$ **then**
11           $A_l^{i+1} = A_l^i \cup \{y \in \mathbb{C} : d(e, y) \leq i, y \notin A_{l2}^i : l2 \in \{1, \ldots, L\}\}$;
12         $i \leftarrow i + 1$;
13 Apply the complement operation to set partition $A_q^{l_q}$ as described in Equation 13;
14 Return $A_1^{l_1}, \ldots, A_L^{l_L}$;

**Algorithm 2**: Construction of the MDC classification model.

### 4.3. Iterative 2D distance

The algorithms presented up to this point consider $p$-dimensional multi-labelled data. In the experiments section, however, we decided to adopt processing $p$-dimensional datasets as combinations of 2-dimensional models. Several 2D "weak" learners can be created from $p$-dimensional data and combined to form a $p$-dimensional classification model. The Random Forest [19] algorithm utilizes a similar concept. It generates several trees with potential reductions on data dimensionality and combines them together to form a general classification model.

Moreover, 2D models are faster to compute and more efficient to store. Not every pairwise combination of attributes has to be explored. Even if they are, it is still more computationally efficient than strict $p$-dimensional dilations (proven in Theorem 4.1). Heuristics can be used to eliminate some of these combinations.

In this section, we introduce a 2-dimensional iterative distance that is efficient when computed using GPUs. The order in which the neighbours are visited in both MkNN and MDC algorithms respects this distance and is shown in Fig. 11, for the upper-right quadrant, where y is the central instance. The 3 remaining quadrants follow the same ordering pattern.

The grey shaded pixels of Fig. 11 can be obtained using the sum of arithmetic progressions, as shown in Eq. 32, where $n$ is the index of the number in the shaded column or line. For instance, number 15 is in the $5^{th}$ line from bottom to top and also in the 5th column from left to right, thus, $n$ would be equal to 5 in this case.

| 15 | 16 | 17 | 18 | 19 | 20 |
|----|----|----|----|----|----|
| 10 | 11 | 12 | 13 | 14 | 19 |
| 6  | 7  | 8  | 9  | 13 | 18 |
| 3  | 4  | 5  | 8  | 12 | 17 |
| 1  | 2  | 4  | 7  | 11 | 16 |
| y  | 1  | 3  | 6  | 10 | 15 |

**Fig. 11.** Upper-right quadrant of the proposed distance metric [58].

$a_1$ is the initial value of the progression and $r$ is its rate which, in this case, both are constants equal to 1.

$$S_n = \frac{n(a_1 + a_n)}{2} \tag{32}$$

where,

$$a_n = a_1 + r(n-1) \tag{33}$$

Finally, after the substitutions, we have Eq. 34.

$$S_n = \frac{n(1+n)}{2} \tag{34}$$

Thereafter, we can state that the numbers to the right of the grey shaded ones are the increments by 1 of their left numbers. So, by fixing the $y$-axis, the $x$ index can be summed to $S_n$, according to $n$, to produce the distance value between two elements. This pattern is repeated with respect to the diagonal line of the table, but the axes must be inverted. Thus, by considering these aspects, the iterative distance formula $d_t(q, w)$ between points $v = (v_1, v_2), w = (w_1, w_2) \in \mathbb{Z}^2$ is shown in Eq. 35.

$$d_t(v, w) = \begin{cases} \frac{x_m(1+x_m)}{2} + y_m & if\ x_m > y_m \\ \frac{y_m(1+y_m)}{2} + x_m & otherwise \end{cases} \tag{35}$$

where,

$$x_m = |v_1 - w_1|, y_m = |v_2 - w_2| \tag{36}$$

### 4.4. Complexity analysis

Theorem 4.1 provides the worst-case complexity and associated analysis of the proposed algorithms assuming that they use the combination of 2-dimensional models to generate a $p$-dimensional classification model. The number of the 2-dimensional models is equal to the number of pairwise combinations of pattern dimensions.

**Theorem 4.1.** *The worst-case step complexity of MkNN and MDC is $O(y_t^2 + x_t^2)$ for 2-dimensions and $O(p^2(y_t^2 + x_t^2))$ for p-dimensional combinations of 2-dimensional classification models in GPU-like environments (Single Instruction Multiple Data paradigm).*

**Proof.** The neighbourhood of each instance in all implementations is checked respecting $d_t$, which accesses each of the neighbouring elements once, respecting the order given by $d_t$. Since classifiers are run in the GPU, each grid instance is theoretically processed in parallel.

The worst case scenario would be dilating an instance in the image at the top-left position $v = (0, 0)$, and its neighbourhood would need to be grown until it covers the entire image. The position $w$ represents the pixel at the right-bottom corner, such that $w = (x_t - 1, y_t - 1)$, where $x_t, y_t$ are the dimensions of the image or grid with respect to the $x$ and $y$ axes. The number of iterations until the image is fully covered is given by the distance function in

Eq. 37.

$$d_t(v, w) = \begin{cases} \frac{|v_2 - w_2|(1 + |v_2 - w_2|)}{2} + |v_1 - w_1|, & \text{if } y_t \geq x_t \\ \frac{|v_1 - w_1|(1 + |v_1 - w_1|)}{2} + |v_2 - w_2|, & \text{if } x_t > y_t \end{cases} \quad (37)$$

Assuming $y_t \geq x_t$, then

$$d_t(v, w) = \frac{|v_2 - w_2|(1 + |v_2 - w_2|)}{2} + |v_1 - w_1|$$

$$d_t(v, w) = \frac{|-y_t + 1|(1 + |-y_t + 1|)}{2} + |-x_t + 1| \quad (38)$$

We can assume that $x_t \geq 1$ and $y_t \geq 1$ to avoid null or empty datasets. In this case, the modulus can be removed, since the result inside them would be positive regardless. $|-y_t + 1|$ outputs values equal or greater than 0. Thus, by assuming that $y_t$ is at least 1, we modify our parenthesis to accommodate this assumption. This is rewritten as $(y_t - 1)$:

$$d_t(v, w) = \frac{(y_t - 1)(1 + (y_t - 1))}{2} + (x_t - 1) \Rightarrow$$

$$d_t(v, w) = \frac{(y_t - 1)(y_t)}{2} + (x_t - 1) \Rightarrow$$

$$d_t(v, w) = \frac{(y_t^2 - y_t)}{2} + (x_t - 1)$$

Thus, $\frac{(y_t^2 - y_t)}{2} + (x_t - 1)$ is the worst-case 2D step complexity if $y_t \geq x_t$. This translates to:

$$O(y_t^2) \quad (39)$$

For $y_t < x_t$ the process is exactly the same, and in this case we obtain:

$$O(x_t^2) \quad (40)$$

Therefore, the worst-case complexity for constructing 2D classification models is $O(y_t^2)$ if $y_t \geq x_t$ or $O(x_t^2)$ otherwise. In other words, the complexity of 2D implementations is $O(y_t^2 + x_t^2)$.

The worst case complexity for the ensemble approach, where 2D dimensional models are combined considering every attribute in $p$-dimensions is:

$$O\left(\binom{p}{2}(y_t^2 + x_t^2)\right) \quad (41)$$

which is the cost for generating a 2D model multiplied by the number of combinations of every two attributes. Expanding the equation:

$$O\left(\frac{p!(y_t^2 + x_t^2)}{(p-2)!}\right)$$

$$O\left(\frac{p(p-1)(p-2)!(y_t^2 + x_t^2)}{(p-2)!}\right)$$

$$O(p(p-1)(y_t^2 + x_t^2))$$

$$O((p^2 - p)(y_t^2 + x_t^2))$$

and finally, we obtain:

$$O(p^2(y_t^2 + x_t^2)) \quad (42)$$

□

As previously discussed, combining 2D models is much better in practice than computing dilations for $p$-dimensions directly. If we apply the expansions in 3 dimensions, for instance, we get a cubic complexity of $O(x_t^3 + y_t^3 + z_t^3)$. Usually, $y_t$ and $x_t$, which represent the size of the grid in 2 dimensions, are substantially greater than the number of dimensions $p$. Let us suppose a hypothetical but frequent scenario whereas $y_t = x_t = z_t = sp$, where $s$ is a scaling factor that relates the size of the grid to the number of dimensions of the problem. The proposed ensemble approach gives

$$p^2((sp)^2 + (sp)^2) = 2s^2 p^4 \quad (43)$$

operations. In the case of expanding the set partitions in $p$-dimensions directly,

$$(sp)^3 + (sp)^3 + (sp)^3 = 3s^3 p^3 \quad (44)$$

operations are required. It is clear that if $s$ is not small, the number of operations in the second case ($3s^3p^3$) is much greater than in the first case ($2s^2p^4$). In fact, if $s$ is greater than $\frac{2p}{3}$, which is the case for every practical classification problem, the number of operations of the latter case is already greater than the former.

In any case, if the number of dimensions $p$ of the problem is equal to 1, which is theoretically the best case, the complexity in the 2D combinations case is $2s^2$ and in the $p$-dimensional case $3s^3$. In the best possible case, it is already clear ($2s^2 < 3s^3$), i.e., it is more efficient to use the 2D combinations approach.

#### 4.4.1. 2D model combinations in practice

Let us suppose that an arbitrary input dataset has 4 dimensions, labelled as $\{d_0, d_1, d_2, d_3\}$, excluding the class label, such as the original UCI Iris dataset. In this case, we have a total of 6 combinations of 2 pairs of features ($d_0$ and $d_1$, $d_0$ and $d_2$, $d_0$ and $d_3$, $d_1$ and $d_2$, $d_1$ and $d_3$, and $d_2$ and $d_3$). Each one of these combinations is a 2D dataset. 2-dimensional models are trained with each of these combinations, and hence each combination produces a respective classification or predictive model. In total, we have 6, 2-dimensional classification models.

Now, let us suppose that we would like to classify instances using these generated predictive models. Each combination produces a vote for the label of an unlabelled instance. These votes can be combined using different paradigms. A simple approach would be to label unlabelled instances with the label that received the most votes. This is exactly what is done in the $p$-dimensional experiments in Section 5.

We consider that some of these 2D models do not produce very accurate votes. Thus, just the votes of the best 2D models are considered, i.e., the ones that achieved the highest accuracies in training and testing with one of the 2D combinations (the training and validation is performed using 10-fold cross validation). The number of 2D models that should have their vote considered was chosen empirically for each experiment. We can fairly say that numbers close to 0 work far better than numbers close to $\binom{p}{2}$, as determined empirically.

The number of combinations is related to the number of features in the dataset, which is not related at all to the number of instances. Iris has 4 features and 150 instances. It is possible to increase the number of instances indefinitely without altering the number of features and vice-versa. Thus, the potential increase in complexity using the 2D combinations would be a vote (which is O(1), since in 2D it is possible to use matrices, and a single position in the matrix is visited), multiplied by the number of combinations. The complexity for generating a 2D model is $O(y_t^2 + x_t^2)$, as previously highlighted. Thus, the complexity for generating 6 of these models is given by the multiplication of this value with the number of combinations, which is given by $\binom{p}{2}$, where $p$ is the number of dimensions (4 for the Iris dataset). The resultant complexity is $O(p^2(y_t^2 + x_t^2))$, which is the worst-case complexity reported in Theorem 4.1.

#### 4.5. 2D model combination theory

Multi-label $p$-dimensional classification is defined as the aggregation of 2 dimensional binary classification models. A multi-label classification, where class labels $l$ belong to $\{1, \ldots, L\}$, is decomposed in $L$ binary problems. For binary problem $l = 1$, the classification models predict if an unlabelled instance belongs to class 1 or not. For binary problem 2, the same follows, the associated models are trained to predict if an unlabelled instance belongs to class 2 or not. This is repeated for all $L$ class labels.

The final class label is given by the aggregation of these $L$ votes, respecting the accuracy of each binary problem, obtained using 10-fold cross validation during the training phase of the algorithm. The votes of classes that obtain the highest accuracies are computed first. Let us suppose we have the following occasion:

$$\begin{aligned}Accuracy : 90.4\% &\leftarrow label : 1 \\ Accuracy : 70.1\% &\leftarrow label : 2 \\ Accuracy : 92.6\% &\leftarrow label : 3\end{aligned} \quad (45)$$

At first, since class 3 has the highest accuracy (92.6%), its votes are considered first in detriment of the other classes. That is, if this classification model labels an unlabelled instance with label 3, then it is predicted to be label 3. If this is not the case (i.e., not label 3), then the instance is tested with the model that has the second highesr accuracy, i.e., 90.4%, and the same procedure is repeated. If it is labelled as 1, then it is said to be label 1. Otherwise, it is labelled as 2. For this procedure to work well, true positive rates are expected to be greater than true negative rates in the binary models. Otherwise, a contrary logic should be used instead. Logical combinations of these votes can be used as well.

For each binary labelling problem, the dataset is decomposed in pairwise combinations of features, as previously discussed. $\binom{p}{2}$ 2-dimensional classification models are produced, and each 2D combination provides a vote for the binary problem.

A total of $\binom{p}{2}$ or less votes are considered for each binary labelling procedure. Some votes can be disregarded according to the accuracy of each 2D model, since they may provide weak or bad votes. At first, all 2D models are ranked according to their accuracy obtained during the 10-fold cross validation process in the training phase. Next, a threshold parameter is set, which indicates how many models should be considered in the voting process. Let us consider the case where there is a total of 22, 2D models (20 combinations of pairwise features), and the threshold indicates that just the top 10 models are trustworthy, then the remaining 12 models are disregarded. In this case, for the given binary problem, the algorithm provides 10 votes, each one indicating whether the instance belongs to a class or not. The majority of the votes indicates the predicted label.

### 4.6. Classification model compression

The classification models produced by the proposed approach are, in theory, sets that contain $p$-dimensional instances. Storage of the model, in practice, can be performed using the said sets or using $p$-dimensional matrices.

#### 4.6.1. Rectangular compression for sets
In the first case, the classification model is given by:

$$A_l^{l_i} = \{(h_1, \ldots, h_p)_1, \ldots, (h_1, \ldots, h_p)_n\} \quad (46)$$

where $p$ represents the dimension of the problem and $n$ represents the total amount of instances in the classification model.

For $p$ dimensions, $pnk$ bits are required to store the model as a set, as demonstrated below:

$$\begin{array}{lcl}\{(h_1)_1, \ldots, (h_1)_n\} & : & nk \text{ bits} \\ \{(h_1, h_2)_1, \ldots, (h_1, h_2)_n\} & : & 2nk \text{ bits} \\ \vdots & : & \vdots \\ \{(h_1, \ldots, h_p)_1, \ldots, (h_1, \ldots, h_p)_n\} & : & pnk \text{ bits}\end{array} \quad (47)$$

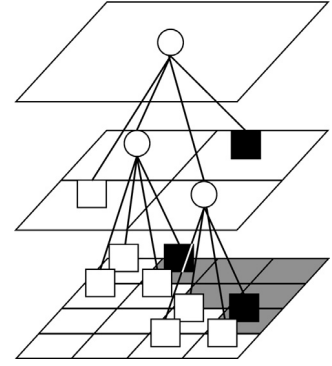

**Fig. 12.** A 2D binary example of a $2^p$-tree.

where $k$ is the amount of bits designed for each variable that, in turn, represent the coordinates of the elements in the set.

However, if the rectangles are stored in a compressed notation (e.g., $(x : x + \theta_x, y : y + \theta_y)$ for two dimensions), where $\theta_j$ represents the size of the rectangle in dimension $j$, $k$ additional bits are required to indicate this extended notation and a further $pk$ additional bits are required to store all $\theta_j$. The required amount of bits is given by:

$$\begin{array}{ll}\{(h_1 : h_1 + \theta_1, h_2 : h_2 + \theta_2)_1 \\ \phantom{\{}(h_1 : h_1 + \theta_1, h_2 : h_2 + \theta_2)_2\} & : \quad 2(4k + k) \text{ bits} \\ \phantom{\{}\vdots & \phantom{:} \quad \vdots \\ \{(h_1 : h_1 + \theta_1, \ldots, h_p : h_p + \theta_p)_1, \ldots, \\ \phantom{\{}(h_1 : h_1 + \theta_1, \ldots, h_p : h_p + \theta_p)_r\} & : \quad r(2pk + k) \text{ bits}\end{array} \quad (48)$$

where $r$ represents the total amount of rectangles.

Thus, a single rectangle of size $(\theta_1, \ldots, \theta_p)$ requires $(2pk + k)$ bits. Following this reasoning, $r(2pk + k)$ bits are required to store $r$ rectangles. In contrast, storing the same $r$ rectangles using Eq. 47, assuming $p$ = dimensional rectangles requires $r(\theta_1 \times \theta_2 \times \ldots \times \theta_p)$ elements, which is the area of the $p$-dimensional rectangles multiplied by their number. This results to a $r(\theta_1 \times \theta_2 \times \ldots \times \theta_p)k$ bits requirement. The equivalence of these two equations is given by:

$$r(2pk + k) = r(\theta_1 \times \theta_2 \times \ldots \times \theta_p)k \quad (50)$$

which is satisfied when $p$ is greater than $(\theta_1 \times \theta_2 \times \ldots \times \theta_p)/2 - 1/2$. This shows that if $p$ is lessthan $(\theta_1 \times \theta_2 \times \ldots \times \theta_p)/2 - 1/2$, the compression approach is much more efficient, which is virtually always the case (very small rectangles being the only expection).

#### 4.6.2. Partition trees for matrices
The second option is to consider images or matrices to store the classification models. In this case, another type of compression can be used. We propose the use of $2^p$-trees to separate the matrix in $p$-dimensional quadrants, i.e., orthants. One of the positive aspects of working with matrices is that most implementations of morphological operators are based on matrices or, for the 2D case, images. Fig. 12 shows a $2^p$-tree representing a 4x4 binary image. Fig. 13 shows the same $2^p$-tree in a flat perspective, where the used orientations are the left-right and top-bottom.

The $2^p$-tree representation consists of dividing the entire matrix in $2^p$ orthants, recursively, until the desired representation precision is reached. In Fig. 12, the image is divided in 4 quadrants at first. The top-right and bottom-left quadrants contain uniform colors, black and white, respectively, and these nodes are promptly turned into leaves. The top-left and bottom-right quadrants are further divided in 4 quadrants each due to the fact that the pixel values in these quadrants are not uniform. The next level of the tree is the deepest level in this case, where the quadrants cannot be divided any more. The leaves of this tree contain values for the binary case (black and white), however it is possible to generalize the representation for grey-scale images, i.e., multi-label classification models, where the leaves of the tree store the respective grey-level (or label) value.

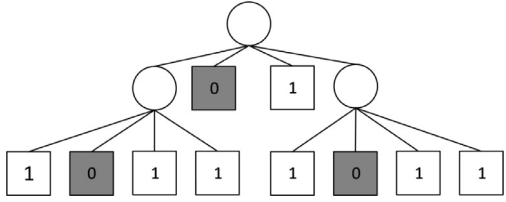

**Fig. 13.** Flat representation of the $2^p$-tree.

It is clear that when the tree branches halt prematurely, a storage compression is achieved. In Fig. 12, two of the nodes of the second level are actually leaves, which were prematurely pruned due to the uniform pattern of the data in their respective quadrants. Thus, the 8 elements of the original image are compressed into 2. The worst-case complexity for reaching any element in the tree is $O(\log_{2^p} n)$, which is very efficient. The image shown in Fig. 12 is written as:

- "XX01X10111011"

in depth-first order, which contains 13 values. 'X' represents a branch or parent node. The original image contains 16 values and the compressed version contains 13. In this case, a compression of 3 elements is obtained. A more expressive compression can be observed in large uniform classification models. The classification models obtained with the proposed algorithms, as shown in the results section, are highly prone to this type of compression, i.e., they are uniform in most regions.

For $p$-dimensional classification matrices, each element in the string representation must represent all possible classes in the problem, i.e., $k$ must respect $2^k \geq L$, where $L$ is the amount of possible distinct classes.

#### 4.6.3. Run length encoding

Finally, a text-compression method, such as run-length encoding [2,7,45] can be applied to both approaches, i.e., the one based on sets and the one utilizing $p$-dimensional matrices, before storing the classification model. The set, as well as the $2^p$-trees, can be written as a single string and therefore text-compression, in general, and run-length encoding, in particular, apply.

Run-length encoding consists of replacing characters repetitions with a unique occurrence of the character along with the the number of times that the character occurrence happens in sequence. The sequence "AAAABBC" can be written as "3A2BC", which requires less memory. Run-length encoding can also be applied along with some pattern recognition approaches, where patterns that occur most frequently are replaced by a predefined set of patterns that requires less memory. In summary, a number of approaches may be used to mitigate the memory burden of the proposed classification model, despite the fact that this is by and large resolved by using the combinations of 2D models instead of the direct $p$-dimensional approach. An appropriate discretization/quantization of the classification space also mitigates the storage requirements.

### 4.7. Parameter selection

The performance of classifiers depends on the chosen set of input parameters. In order to fairly compare the performance of morphological classifiers to other classification algorithms and to visualize the obtained results, we performed two separate sets of experiments. At first, experiments with 2 dimensional datasets were performed. That is, two $p$-dimensional datasets were reduced to 2 dimensions, and the parameters of each classifier were tuned using an evolutionary algorithm (EA) [46].

Parameters were optimized using two methods. The first approach consists of using EA for tuning of numerical parameters, while categorical attributes are manually configured. The second approach uses Auto-Weka [17], which determines the best suited algorithm from the Weka [18] machine learning framework and an appropriate set of parameters, including a subset of the dataset attributes, if it leads to a performance improvement.

#### 4.7.1. Evolutionary algorithm

The proposed EA consists of a combination of an elitist strategy, mutation and cross-over. The size of the population carried on to the next generation was set to 20. At first, 20 individuals are randomly generated. In each generation, a single individual is created; this receives randomly selected parameters of two randomly selected parents (cross-over). After that, mutation is performed on the selected parameters. This individual has its objective function computed to confirm whether it may enter in the top 20 members of the population. If this is true, then the individual with the smallest objective function value among the top 20 is removed. The objective function is defined as the accuracy of the classifier on 10-fold cross validation. Algorithm 3 illustrates the overall process of the EA.

**Data**: $\Theta$ is the population at each generation, $\Psi()$ is a function that generates a random number $\in [0, 1]$, $\varepsilon$ represents individuals and $\Pi$ represents a maximal value of each parameter, chosen according to the classification algorithm.

1 **begin**
2    $w \leftarrow 0$;
3    $\Theta \leftarrow$ randomly generate 20 individuals;
4    **while** *the target time has not been reached* **do**
5      $\varepsilon_1 \leftarrow$ take the $(40\Psi() \% 20)^{th}$ best fit individual in $\Theta$;
6      $\varepsilon_2 \leftarrow$ take the $(40\Psi() \% 20)^{th}$ best fit individual in $\Theta$;
7      $\varepsilon_{new} \leftarrow$ randomly fuse the parameters of $\varepsilon_1$ and $\varepsilon_2$ (e.g., take parameter 1 from $\varepsilon_2$, parameter 2 from $\varepsilon_1$, parameter 3 from $\varepsilon_1$, etc, where this choice is completely at random until all parameters are populated);
8      $\varepsilon_{new} \leftarrow$ perform a mutation on $\varepsilon_{new}$ if $\Psi() < 0.6$, i.e., for each parameter $s$, if $\Psi() < 0.6$, then the parameter $s = s + \frac{\Psi() \Pi_s - \frac{\Psi() \Pi_s}{2}}{\frac{w}{50}}$;
9      **if** *the objective function of $\varepsilon_{new}$ is higher than of any individual in $\Theta$* **then**
10       Take out the least fit individual from $\Theta$ and insert $\varepsilon_{new}$;
11     $w \leftarrow w + 1$;

**Algorithm 3:** Evolutionary algorithm to select the numerical set of parameters for the classification algorithms.

Each EA run was limited a total of 12 hours. This was also the case for Auto-Weka. Since MkNN and MDC were faster than the algorithms in Weka, for these cases, the genetic EA was set to converge in 6 hours.

#### 4.7.2. Auto-Weka

As opposed to EA, the results of Auto-Weka were not manually tweaked. Although Auto-Weka performs a 10-fold cross validation internally, it presents its indexes (accuracy, true positives, etc) tested on the entire training set. Therefore, the reported results are somewhat unfair with regards to the first methodology (EA plus manual tweaking), where the results are presented as tested on the 10 partitions of the 10-fold cross validation. The authors of Auto-Weka reported that this was defined due to how Weka works internally [17,47]. If the final results reported by Auto-Weka were derived from cross validation alone, the reported indexes would be lower in comparison.

**Table 1**
Datasets used in the 2d experiments.

| Dataset | Selected Attributes | Number of Instances |
|---|---|---|
| Iris | sepallength, petallength | 100 (50 – 50) |
| Diabetes | plas, insu | 768 (500 - 268) |

## 5. Experimental results

Weka [18], a Java based machine learning framework, was used to run and test the classification algorithms apart from MDC and MkNN. MkNN and MDC were implemented in CUDA/C/C++ and their source code is available at [48]. All the implementations proposed in this work conform to the use of the complement approach shown in Theorem 3.1 due to the previously described benefits.

### 5.1. 2-dimensional experiments

This section addresses the details of datasets used in our experiments as well as the performance of the classifiers, visual results and comparisons. The parameters were selected, as described in Section 4.7. We performed experiments with 2-dimensional subspaces of $p$-dimensional datasets. In this regard, predictive models can be visualized and analysed as an image. In the next section, experiments with $p$-dimensional datasets are performed.

The first implementation of the proposed classifiers considers input datasets as multisets, that is, repeated instances, i.e., instances where the feature values are identical, influence the set expansion. The second implementation excludes these repetitions. The subscript *Rep* in the name of the algorithms indicates the version that uses the multiset concept. As a remark, combining 2D models is more efficient than using $p$-dimensional dilations, as shown in Section 4.4.

#### 5.1.1. Used datasets

Two reduced versions of datasets from the UCI repository [49] were considered in this set of experiments. The first one is the widely known *Iris* dataset. The number of attributes of this dataset was reduced to 2, i.e., attributes *sepallength* and *petallength* were randomly selected and the remaining features were excluded, except for the class label. Instances whose classes were equal to *Iris-setosa* were also removed from the dataset in order to transform this dataset into a binary classification problem.

The second dataset was *Diabetes*, and the selected attributes were *plas* and *insu*. The original dataset already consists of a binary classification problem and thus no instances were removed. Table 1 illustrates this information and the total number of instances of each dataset.

#### 5.1.2. Iris dataset

The results obtained with the Iris dataset and each algorithm (optimized with EA along with manual adjustments) as well as the Auto-Weka approach, which is highlighted in *italic* - refer to Section 4.7 for more information, are shown in Table 2.

The parameters of the morphological classifiers were also automatically selected using the EA approach (Section 4.7). Among

**Table 2**
Performance comparison on the Iris dataset.

| Classifier | Acc (%) | TP (%) | TN (%) |
|---|---|---|---|
| Multilayer Perceptron [50] | **95** | 92 | 98 |
| RBF Network [21] | 94 | **94** | 94 |
| Conjunctive Rule [50] | 94 | 92 | 96 |
| SPegasos [50] | 93 | 92 | 94 |
| k-NN (IBk) [13] | 93 | **94** | 92 |
| SVM [51] | 93 | 92 | 94 |
| SMO [25] | 93 | 92 | 94 |
| *Auto-Weka* [17] | 93 | 92 | 94 |
| REPTree [50] | 92 | 88 | 96 |
| Random Forest [19] | 91 | 92 | 90 |
| Decision Table [52] | 91 | 88 | 94 |
| C4.5 (J48Graft) [50] | 91 | 88 | 94 |
| C4.5 (J48) [53] | 91 | **94** | 88 |
| Bayes Net [54] | 91 | 88 | 94 |
| Naive Bayes [55] | 91 | 88 | 94 |
| Hoeffding Tree [56] | 87 | 88 | 86 |
| **MkNN** | **95** | **98** | 92 |
| **MkNN**$_{Rep}$ | 94 | **96** | 92 |
| **MDC** | **95** | **100** | 90 |
| **MDC**$_{Rep}$ | **95** | **98** | 92 |

**Rep** stands for the version of the algorithm that considers instance repetitions (multiset).

The algorithms highlighted in **bold** are the ones proposed in this work.

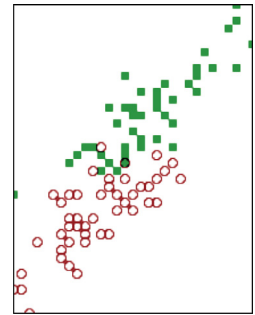

**Fig. 14.** Preprocessed Iris dataset.

the classifiers in comparison, Auto-Weka selected Logistic Model Trees (LMT) [57] as the best suited approach. The results were obtained using 10-fold cross validation (except for Auto-Weka, refer to Section 4.7).

Apart from morphological classifiers, Multilayer Perceptrons achieved the highest accuracy, i.e., 95%. Three implementations of the proposed algorithms also achieved 95% of accuracy with this dataset (MkNN, MDC and MDC$_{Rep}$). In most cases, MkNN and MDC produce high true positive rates, usually higher than the remaining evaluated classification algorithms while maintaining high accuracy (this is more evident with the MDC algorithm).

Fig. 14 shows the utilised 2D Iris dataset after the preprocessing described in the beginning of Section 4. The classes that are equal to 1 are shown as red circles while the ones that are equal to 2 are depicted as green squares.

10-fold cross validation separates the dataset in 10 partitions. Next, 10 iterations follow (the folds). In the first iteration, a classification model is created trained on $10 - 1$ of the partitions and tested on the remaining 1. This process is repeated 9 more times while always switching the test partition to a partition that has not been previously used as test set.

Fig. 15 shows one of the classification models of the 10 iterations (or folds) of the Iris dataset shown in Fig. 14. These images are classification models $A_1^{I_1}$ of the implementations (a) MkNN, (b) MkNN$_{Rep}$, (c) MDC and (d) MDC$_{Rep}$, respectively. The 10 generated classification models for each fold can be found in [48].

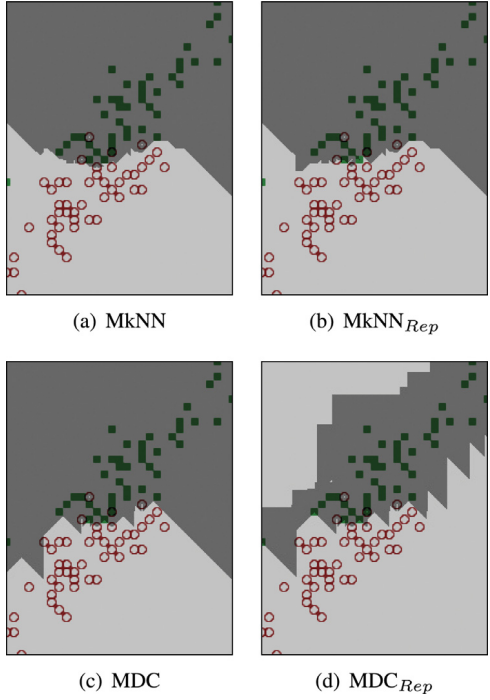

**Fig. 15.** Generated classification models on Iris.

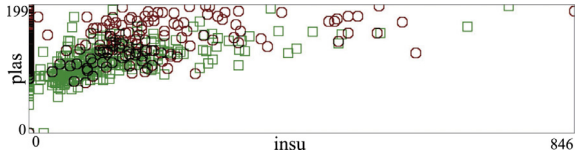

**Fig. 16.** Preprocessed Diabetes dataset.

The dark grey area in Fig. 15 represents what does not belong to $A_1^{I_1}$, while the light grey represents instances that belong to $A_1^{I_1}$. Assuming that repeated instances are discarded and conceiving the fact that this dataset almost does not contain repeated instances, then the larger the area where the light gray intersects instances labelled as 1 (red circles) and excludes the ones labelled as 2 (green squares), the higher the accuracy of the model.

### 5.1.3. Diabetes dataset

Fig. 16 shows the 2D Diabetes dataset. Fig. 17, on the other hand, shows the classification models of one fold of each proposed algorithm. In the case of $MDC_{Rep}$, the orientation parameter $\beta$ chosen by the EA was equal in all directions but the left. Therefore, the dilator clearly grows the structure in the remaining 3 directions. In the case of MDC, $\beta$ restricted the growth downwards.

Table 3 shows the obtained results on the 2D Diabetes dataset. In this case, $MkNN_{Rep}$ and SMO outperformed the remaining algorithms, and tied on the accuracy. However, MDC still produces a high true positive rate. However, the accuracy obtained with MDC was 71.61%, which is considerably less than the 74.86% obtained by SMO. Auto-Weka, in this case, also selected LMT as the best classifier.

### 5.2. P-Dimensional Experiments

In this subsection, experiments and comparisons are performed with p-dimensional binary and multi-label datasets. We considered 8 datasets from the UCI public repository. The proposed p-dimensional approach uses a combination for 2-dimensional clas-

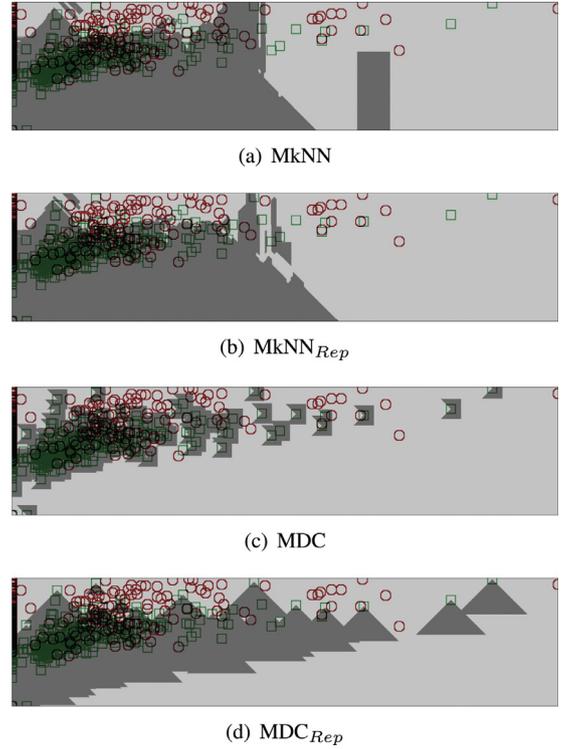

**Fig. 17.** Generated classification models on Diabetes dataset.

**Table 3**
Performance comparison on the Diabetes dataset.

| Classifier | Acc (%) | TP (%) | TN (%) |
|---|---|---|---|
| SMO [25] | **74.86** | **91.6** | 43.7 |
| SPegasos [50] | 74.60 | 90.6 | 44.8 |
| Multilayer Perceptron [50] | 74.60 | 88.2 | 49.3 |
| SVM [51] | 74.60 | 89.8 | 46.3 |
| *Auto-Weka* [17] | 74.47 | 88 | 49.3 |
| Hoeffding Tree [56] | 74.08 | 89.4 | 45.5 |
| RBF Network [21] | 73.82 | 90.4 | 42.9 |
| Naive Bayes [55] | 73.69 | 89.8 | 43.7 |
| k-NN (IBk) [13] | 73.43 | 84.8 | 52.6 |
| REPTree [50] | 73.04 | 86.6 | 47.8 |
| C4.5 (J48Graft) [50] | 73.04 | 88.8 | 43.7 |
| C4.5 (J48) [53] | 73.04 | 88.8 | 43.7 |
| Decision Table [52] | 72.65 | 90.6 | 39.2 |
| Conjunctive Rule [50] | 72.52 | 81.4 | 56 |
| Bayes Net [54] | 71.35 | 83.6 | 48.5 |
| Random Forest [19] | 68.48 | 80.4 | 46.3 |
| **MkNN** | 72.65 | 88.6 | 46.2 |
| **MkNN**$_{Rep}$ | **74.86** | 88.6 | 49.2 |
| **MDC** | 71.61 | **91.4** | 37.7 |
| **MDC**$_{Rep}$ | 73.43 | 89 | 44.4 |

**Rep** stands for the version of the algorithm that considers repeated instances (multiset).
The algorithms highlighted in **bold** are the ones proposed in this work.

sification models, due to previously described advantages (see Theorem 4.1).

Each pairwise combination of features generates a binary predictive model. The total number of models is given by $\binom{p}{2}$, where $p$ represents the number of features or dimensions of the dataset. An unlabelled instance is classified using the votes of all $\binom{p}{2}$ models, where each vote corresponds to a binary vote, belonging or not to a class.

Multi-label datasets were separated in L binary problems, where L is the number of labels of the dataset. That is, binary classifiers were trained for each of the L classes. Let us consider

**Table 4**
Accuracies in percentage obtained with each dataset and classification algorithm.

| Classifiers | Iris | Diabetes | Liver Disorders | Tae | Column 2C | Haberman | Heart Statlog | Breast W |
|---|---|---|---|---|---|---|---|---|
| RandomForest | 95.33 | **77.78** | 61.15 | 68.87 | 81.48 | 69.28 | 81.48 | 96.70 |
| Multilayer Perceptron | 97.33 | 75.67 | 65.07 | 54.96 | 78.51 | 69.28 | 78.51 | 94.99 |
| RBFNetwork | 95.33 | 75.39 | **66.95** | 52.98 | 84.07 | 73.85 | 84.07 | 95.85 |
| Conjunctive Rule | 66.66 | 71.87 | 60.86 | 37.74 | 74.07 | 73.52 | 74.07 | 91.70 |
| kNN | 96.66 | 74.73 | 57.68 | 62.25 | 81.11 | 73.85 | 81.11 | 96.70 |
| SVM | 96.66 | 77.47 | 56.81 | 54.30 | 66.66 | 75.49 | 66.66 | 96.85 |
| SMO | 96.66 | 77.34 | 64.63 | 58.27 | 84.07 | 73.85 | 84.07 | 97.28 |
| REPTree | 94.00 | 75.26 | 61.44 | 54.30 | 78.51 | 73.52 | 78.51 | 94.27 |
| DecisionTable | 92.66 | 70.18 | 63.47 | 47.01 | **84.81** | 72.54 | **84.81** | 95.27 |
| J48Graft | 94.66 | 73.69 | 60.28 | 60.26 | 77.07 | 73.20 | 77.07 | 94.84 |
| J48 | 96.00 | 73.82 | 59.13 | 59.60 | 76.66 | 72.87 | 76.66 | 94.56 |
| NaiveBayes | 96.00 | 76.30 | 64.05 | 54.30 | 83.70 | **76.14** | 83.70 | 97.13 |
| BayesNet | 94.00 | 74.34 | 61.44 | 47.01 | 81.11 | 72.54 | 81.11 | 97.13 |
| HoeffdingTree | 95.33 | 76.17 | 63.76 | 53.64 | 83.30 | 74.83 | 83.30 | 95.99 |
| **MDC**$_{Rep}$ | **98.00** | 75.39 | 65.79 | 66.66 | 81.29 | 75.81 | 84.07 | **97.42** |
| **MkNN**$_{Rep}$ | **98.00** | 75.13 | 66.37 | 72.46 | 81.93 | 76.14 | 84.81 | 96.99 |
| **MDC**$_{Rep}\star$ | 4.62 | 0.39 | 3.88 | 11.98 | 1.64 | 2.61 | 4.42 | 1.76 |
| **MkNN**$_{Rep}\star$ | 4.77 | 0.34 | 4.40 | 18.87 | 2.42 | 2.64 | 5.30 | 1.41 |

$\star$ represents the percentage improvement in comparison to the average accuracy of all state-of-the-art classifiers.
MDC and MkNN refer to the two proposed classifiers.

**Table 5**
Training and testing times (s) obtained with MDC$_{Rep}$ and MkNN$_{Rep}$ in seconds.

| Datasets | MDC Train Time (s) | MDC Test Time (s) | MkNN Train Time (s) | MkNN Test Time (s) |
|---|---|---|---|---|
| Iris | 0.1223382649 | 0.0000030028 | 0.1594423040 | 0.0000009341 |
| Diabetes | 6.4961881040 | 0.0004020527 | 14.2825142510 | 0.0002412080 |
| Liver Disorders | 1.8445822000 | 0.0000200820 | 0.9116863500 | 0.0000295500 |
| Tae | 3.6113764500 | 0.0000900310 | 3.6986199000 | 0.0006007500 |
| Column 2C | 2.9733453099 | 0.0000601520 | 6.1188321300 | 0.0000798789 |
| Haberman | 0.8608457292 | 0.0000100440 | 0.8368868598 | 0.0000100000 |
| Heart Statlog | 9.3438252208 | 0.0001833519 | 13.3958921399 | 0.0023547693 |
| Breast W | 5.1447441331 | 0.0003430842 | 13.9826192997 | 0.0003608940 |

a three-class problem (i.e., classes 1, 2 and 3). For class 1, the instances in the dataset either belong or not to class 1. Thus, this constitutes the binary problem for class 1. A further two binary problems are produced for classes 2 and 3.

Next, for the binary problem of class 1, the approach follows as previously described. $\binom{p}{2}$ binary classification models are generated for the class 1 problem. Each model provides a vote for an unlabelled $p$-dimensional instance. In total, we have $\binom{p}{2}$ or less votes (depending on the significance of the model/vote), which decide whether the instance belongs to class 1 or not. The procedure is repeated for the other two classes, i.e., class 2 and class 3. At the end, each binary problem (class 1, class 2 and class 3) provides a vote for each one of the class labels. The majority of votes decides the predicted label of the unlabelled instance. Section 4.5 describes this decomposition logic in more detail.

Table 4 contains the accuracies obtained with each classifier for each dataset. Table 5 shows the training and testing times of MDC and MkNN. Training times include the training phase and the time required for building the 2D models for all pairwise feature combinations as well as the multi-label training, as applicable for the dataset, considering 9/10ths of the original dataset. Test times represent the time required for testing 1/10th of the dataset (one fold of the 10-fold cross validation). Although training phases require additional time, testing times are remarkably fast due to the nature of the classification model, which are matrices whose elements are accessed with O(1) complexity in the worst case.

## 6. Conclusion

This work proposes a new category of classification algorithms called Morphological Classifiers (MC). MCs are an aggregation of mathematical morphology and supervised learning. The classifiers rely on the fact that datasets may contain well defined shapes and structures, which naturally come under the remit of mathematical morphology and could be a positive aspect to explore. In addition to establishing the theoretical framework of MCs, we propose two GPU implementation approaches.

This work is the first to propose such a type of classifier, which was shown to outperform some well established classifiers in the performed experiments. Experiments presented herein focused on proposing and evaluating 2-dimensional models and further aggregating them to form $p$-dimensional classification models. MDC and MkNN classifiers tied or outperformed 14 well established classifiers in 5 out of 8 $p$-dimensional UCI datasets. In all of these cases, the proposed classifiers obtained accuracies that were higher than the average accuracy across all classifiers.

Although the proposed theoretical framework and algorithms readily work for $p$-dimensional datasets, we proved that it is faster, in terms of required steps in the worst-case scenario, to generate 2-dimensional classification models and to combine them, rather than working directly on $p$-dimensional grids and creating $p$-dimensional classification models using $p$-dimensional dilations. These 2-dimensional models are combined using a voting scheme, as an ensemble classifier.

The complexity for checking whether an instance belongs to a certain class using matrices is $O\binom{p}{2}$ in the worst case, where $p$ stands for the number of features or dimensions of the dataset. This is a very fast instance classification that does not relate to the number of instances in the dataset. The classification time is potentially faster than decision-tree based classification times, which are already fairly fast in comparison to other classification paradigms.

Drawbacks include the fact that training morphological classifiers may be slow, particularly when dilations are performed in

É.O. Rodrigues et al. / Pattern Recognition 84 (2018) 82–96

$p$-dimensions directly, for instance. However, due to the nature of the operations performed, such classifiers are amenable to parallelization. Moreover, in a fashion similar to the training of deep learning networks, training times may not be considered a major disadvantage, as the actual classification process is instantaneous and classification models demonstrate high accuracy rates. As a final remark, the code for the proposed classifiers in CUDA/C/C++ is available as open-source at [48] along with the utlized datasets.

### 6.1. Future work and discussion

Other opportunities to explore in regards to morphological classifiers include introducing more rules for the expansion process of MDC. The current MDC proposal is fairly naive, and this simplicity certainly introduces more classification errors. Growing in a fractal fashion was also evaluated and produced acceptable results. However, the fractal approach was too simple and did not outperform most classifiers. However, the idea of exploring fractal information in datasets and using it in classification problems has not been fully explored. The morphological classifiers framework allows for this possibility.

Another direction for further research is to explore algorithms that are dilated in $p$-dimensions using $p$-dimensional morphological operations such as $p$-dimensional dilations, conditional thickening, etc. We performed a $p$-dimensional dilation experiment with the Iris dataset, and the dilator algorithm obtained the same accuracy as the one obtained with the 2-dimensional models combination approach, shown in Table 4. Direct $p$-dimensional approaches are possible but require more memory and processing power. Thus, future research may focus on reducing this burden by using approximations or real-time functions that generate the classification model in classification time, hence enabling $p$-dimensional morphological approaches to be performed in real time for a broad class of datasets.

Furthermore, data compression techniques can be applied to reduce the burden of memory requirements. We intend to deeply explore the application of such techniques and to propose efficient ways of storing classification models for big datasets in future work. It is necessary to highlight, however, that the 2D combination approach works well even with reasonably large datasets, as the memory and processing burdens are mitigated. Testing and training times using the combination approach can be found in Table 5. Morphological functions could also be used. In this case, it is not necessary to store a discretized classification space. These functions could define whether an instance belongs to a certain area, and hence, to a class. The key question here is whether it is viable to merge different functions and to construct such models during the classification process. This approach will also be further explored in future research.